\pdfoutput=1

\documentclass[11pt]{article}

\usepackage{acl}
\usepackage{stfloats}

\usepackage{times}
\usepackage{latexsym}
\usepackage{booktabs}
\usepackage{tabularx}
\usepackage[utf8]{inputenc}
\usepackage{microtype}
\usepackage{inconsolata}
\usepackage[T1]{fontenc}
\usepackage{textcomp}
\usepackage{tikz}
\usepackage{graphicx}

\usepackage{titlesec}

\titlespacing*{\section}{0pt}{1.5ex plus .2ex minus .2ex}{0.8ex}

\newcommand{\ignore}[1]{}

\title{Learning from Child-directed Speech in Two-language Scenarios: A French-English Case Study}

\author{Liel Binyamin, Elior Sulem\\
  Faculty of Computer and Information Science, Institute for Applied AI Research\\
  Data Science Research Center\\
  Ben-Gurion University of the Negev \\
  \texttt{lielbin@post.bgu.ac.il, eliorsu@bgu.ac.il} 
 }

\begin{document}
\maketitle

\begin{abstract}
Research on developmentally plausible language models has so far centered on English, leaving open questions about multilingual settings. We present a systematic study of compact models by extending BabyBERTa to English–French scenarios under strictly size-matched data conditions, addressing {\it monolingual}, {\it bilingual}, and {\it cross-lingual} settings. Our design contrasts two corpus types: (i) child-directed speech ($\approx$2.5M tokens), following BabyBERTa and related work, and (ii) multi-domain corpora ($\approx$10M tokens), extending the BabyLM framework to French. To support fair evaluation, we also introduce new resources: French versions of QAMR and QASRL, and an English and French multi-domain corpus.
We evaluate the models on both syntactic and semantic tasks, comparing with Wikipedia-only training. Results reveal context-dependent effects: training on Wikipedia consistently favors semantic tasks, while child-directed speech improves grammatical judgments in monolingual settings. Bilingual pretraining yields notable gains for textual entailment, disproportionately benefiting French. Importantly, the same relative patterns are observed across BabyBERTa, RoBERTa, and LTG-BERT, indicating consistent trends across the tested architectures.\footnote{Code and data are available at \url{https://github.com/NLU-BGU/Compact-Multilingual-LMs}.}
\end{abstract}

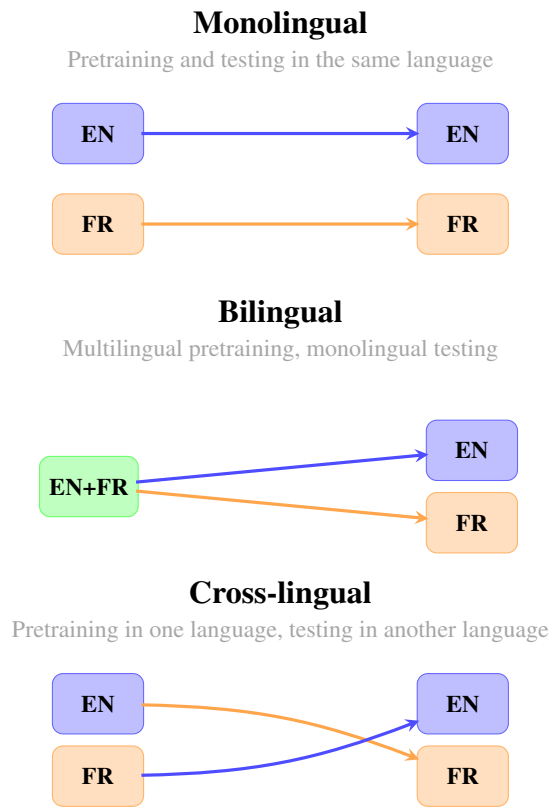
\begin{figure}[t]
\centering
\begin{tikzpicture}[
    scale=1.2,
    node distance=1.5cm,
    lang/.style={draw, rounded corners=4pt, minimum width=1.2cm, minimum height=0.8cm, font=\small\bfseries},
    arrow/.style={->, very thick, >=stealth, shorten >=-1pt, shorten <=-1pt},
    en/.style={lang, fill=blue!25, draw=blue!60},
    fr/.style={lang, fill=orange!25, draw=orange!60},
    enfr/.style={lang, fill=green!25, draw=green!60},
    title/.style={font=\large\bfseries},
    subtitle/.style={font=\small, text=gray!70}
]

\node[title] (mono-title) at (0, 3) {Monolingual};
\node[subtitle] (mono-sub) at (0, 2.6) {Pretraining and testing in the same language};
\node[en] (mono-en-pre) at (-2, 1.8) {EN};
\node[en] (mono-en-fine) at ( 2, 1.8) {EN};
\node[fr] (mono-fr-pre) at (-2, 0.8) {FR};
\node[fr] (mono-fr-fine) at ( 2, 0.8) {FR};
\draw[arrow, blue!70]   (mono-en-pre) -- (mono-en-fine);
\draw[arrow, orange!70] (mono-fr-pre) -- (mono-fr-fine);

\node[title] (bi-title) at (0, -0.2) {Bilingual};
\node[subtitle] (bi-sub) at (0, -0.6) {Multilingual pretraining, monolingual testing};
\node[enfr] (bi-pre)     at (-2.1, -2.1) {EN+FR};
\node[en]   (bi-en-test) at ( 2.1, -1.7) {EN};
\node[fr]   (bi-fr-test) at ( 2.1, -2.5) {FR};
\draw[arrow, blue!70]   (bi-pre) -- (bi-en-test);
\draw[arrow, orange!70] (bi-pre) -- (bi-fr-test);

\node[title] (cross-title) at (0, -3.3) {Cross-lingual};
\node[subtitle] (cross-sub) at (0, -3.7) {Pretraining in one language, testing in another language};
\node[en] (cross-en-pre)   at (-2, -4.5) {EN};
\node[fr] (cross-fr-pre)   at (-2, -5.3) {FR};
\node[en] (cross-en-test)  at ( 2, -4.5) {EN};
\node[fr] (cross-fr-test)  at ( 2, -5.3) {FR};
\draw[arrow, orange!70, bend left=10]  (cross-en-pre) to (cross-fr-test);
\draw[arrow, blue!70,   bend right=10] (cross-fr-pre) to (cross-en-test);

\end{tikzpicture}
\caption{Monolingual, bilingual, and cross-lingual settings. Left column shows the \emph{pretraining} language(s); right column shows the \emph{testing} language, and these labels apply to all three settings. For downstream QA/TE experiments, fine-tuning is performed in the same language as testing.}
\label{fig:language-strategies}
\end{figure}

\section{Introduction}

Training efficient language models with limited computational resources presents a fundamental challenge in natural language processing. While large-scale models have achieved impressive performance, their computational demands—requiring massive datasets and substantial processing power—make them inaccessible for many research contexts and real-world applications. This challenge is particularly acute for multilingual settings, where resource constraints are compounded by the need to support multiple languages. Beyond practical concerns, efficient language model training offers insights into language acquisition: if models can achieve linguistic competence with human-scale data, they provide controlled empirical settings that can inform future theory-driven work on child language acquisition.

Recent work has demonstrated promising approaches to this challenge. The BabyBERTa model \cite{huebner-etal-2021-babyberta} showed that compact architectures trained on just 5 million words of \textit{child-directed speech} could match larger models' grammatical proficiency while dramatically reducing computational costs. The BabyLM Challenge \cite{warstadt-etal-2023-findings} further established that developmentally plausible training corpora—approximating the linguistic input children receive—can produce models with strong capabilities under significant resource constraints. These studies validate that carefully curated, human-scale datasets enable efficient language model training.

However, gaps remain. First, these foundational studies have focused mainly on monolingual English settings, leaving cross-linguistic validity underexplored. Second, evaluations of models exclusively trained on child-directed speech have concentrated primarily on grammatical competence, with limited investigation of semantic understanding through tasks like question-answering and textual entailment. Third, while some recent work has explored bilingual scenarios \cite{yadavalli-etal-2023-slabert,shen-etal-2024-bambino}, these studies employ sequential pretraining (L1 then L2) and evaluate only in English, rather than examining simultaneous bilingual exposure and balanced cross-linguistic evaluation. These limitations prevent understanding whether efficiency gains generalize across languages and whether small models can achieve deeper semantic understanding.

We address these gaps through systematic investigation of training in two-language scenarios with BabyBERTa, focusing on English and French. We examine three experimental configurations (Figure~\ref{fig:language-strategies}): (1) \textit{monolingual}—comparing monolingual pretraining in English and French with matched data; (2) \textit{bilingual}—simultaneous exposure to both languages during pretraining; (3) \textit{cross-lingual}—pretraining in one language, evaluation in another. Our key methodological innovation is establishing fair cross-linguistic comparisons through two types of corpora: 
(1) \textbf{child-directed speech corpora} ($\approx$2.5M words, constrained by French CHILDES availability), representing developmentally realistic input, alongside matched Wikipedia-only corpora for comparison; and (2) \textbf{multi-domain corpora} ($\approx$10M words, following the BabyLM framework but adapting it to an English-French comparison), representing broader and more heterogeneous input. 
This reframing emphasizes the nature of the training data rather than size alone, while still maintaining comparability across scales. We also introduce French versions of the QAMR \cite{michael-etal-2018-crowdsourcing} and QASRL \cite{he-etal-2015-question} evaluation datasets, enabling balanced cross-linguistic testing.

Our evaluation moves beyond grammar-focused assessments to semantic understanding tasks: question-answering (SQuAD \cite{rajpurkar-etal-2016-squad}, QASRL \cite{he-etal-2015-question}, QAMR \cite{michael-etal-2018-crowdsourcing}), textual entailment \cite[XNLI,][]{conneau-etal-2018-xnli}, and grammatical competence \cite[CLAMS,][]{mueller-etal-2020-cross}. The choice of including question-answering is motivated by natural alignment with our training data—CHILDES provides abundant conversational Q\&A interactions (179,953 in English, 132,133 in French), making question-answering evaluation well-suited to child-directed speech pretraining. We establish strong baselines through comparisons with RoBERTa models trained on our configurations and pretrained models (RoBERTa-base \cite{zhuang-etal-2021-robustly}, CamemBERT \cite{martin-etal-2020-camembert}). We also conduct supplementary analyses with LTG-BERT \cite{samuel-etal-2023-trained}, on which the top BabyLM Challenge performer is based, and on T5-tiny \cite{mueller-linzen-2023-plant} to examine whether the observed patterns extend beyond a single architecture.

Our results demonstrate that (1) bilingual training disproportionately benefits textual entailment in the weaker language, (2) French benefits from exposure to child-directed speech when combined with Wikipedia training, and (3) small-scale multilingual models can acquire meaningful semantic capabilities beyond grammatical competence, with consistent patterns observed across the tested architectures considered here.

\section{Related Work}

\paragraph{Child-Directed Speech Models}
The BabyBERTa model \cite{huebner-etal-2021-babyberta} demonstrated that child-directed speech (CDS) could train efficient language models with strong grammatical proficiency despite limited data and computational resources. A compact RoBERTa architecture trained on just 5 million words of CDS achieved comparable grammatical competence to much larger models, establishing the feasibility of developmentally plausible language model training. \citet{mueller-linzen-2023-plant} further demonstrated that pre-training on child-directed speech induces hierarchical syntactic biases in small encoder-decoder models with order-of-magnitude less data than typical web text or Wikipedia corpora, suggesting greater data efficiency in cognitively plausible acquisition settings. However, these works focused on English and on grammatical competence rather than semantic understanding tasks. Extensions of these studies to other languages include \citet{bunzeck-etal-2025-construction} for German and \citet{gelboim-sulem-2025-tafberta} for Hebrew. \citet{yang-etal-2023-bootstrapping} evaluated models trained on English CDS on downstream tasks, covering Semantic Role Labeling, QASRL, and QAMR.

\paragraph{BabyLM Challenge and Multidomain Training}
The BabyLM Challenge \cite{warstadt-etal-2023-findings} extended efficient language model training by promoting development of small-scale models optimized for language acquisition scenarios using multidomain corpora. The challenge, which includes both grammatical \citep[BLiMP,][]{warstadt-etal-2020-blimp-benchmark} and downstream task \citep[GLUE,][]{wang-etal-2018-glue}  evaluation, demonstrated that various architectural innovations could enhance learning efficiency within constrained data settings, using diverse developmentally plausible sources beyond child-directed speech alone.
ELC-BERT \cite{georges-gabriel-charpentier-samuel-2023-layers} emerged as the winning model, representing a paradigm shift toward architectural optimization over data scaling. ELC-BERT is based on the LTG-BERT model \citep{samuel-etal-2023-trained}. Training on the British National Corpus (100 million words), LTG-BERT incorporated four key innovations: NormFormer normalization, GEGLU activation functions, disentangled attention, and span masking. Despite using significantly smaller corpora than standard models, LTG-BERT demonstrated that careful architectural design can achieve strong language understanding with developmentally plausible data amounts. \citet{WILCOX2025104650} found that ELC-BERT and LTG-BERT achieve comparable performance, concluding that it is the innovations incorporated in LTG-BERT that led to the win.

Our work extends this line of research by investigating whether efficiency gains observed in monolingual settings transfer to two-language scenarios, focusing on English and French. We consider both CDS-only and multidomain setups and include extractive question answering of various levels of complexity in the evaluation.
While RoBERTa serves as our main baseline, we also include LTG-BERT as a competitive reference point in the case of multidomain training.

\paragraph{Multilingual and Cross-Lingual Training in Small Models}
Several studies have explored cross-lingual transfer in neural language models with different focuses than our work. SLABERT \cite{yadavalli-etal-2023-slabert} investigated cross-lingual transfer by pretraining BERT models on CDS data in multiple languages before transitioning to Adult-Directed Speech in English, evaluating transfer effects using grammatical judgment tasks (BLiMP). Research on second language acquisition in neural models \cite{oba2023secondlanguageacquisitionneural} followed a three-step process of L1 pretraining, L2 continual training, and grammatical evaluation, while work on modeling nonnative sentence processing \cite{aoyama-schneider-2024-modeling} examined how L1 influences L2 processing through reading-time analysis. BAMBINO-LM \cite{shen-etal-2024-bambino} explored bilingual pretraining through continuous learning, where a small model was incrementally trained on a second language with guidance from a larger parent model, simulating human bilingual acquisition through sequential L1-to-L2 training.

These studies differ from our work in key aspects. All employ sequential pretraining paradigms (L1 then L2) rather than examining simultaneous bilingual exposure as occurs in natural child bilingual acquisition. They focus primarily on grammatical competence evaluation and conduct evaluations exclusively in English (except BAMBINO-LM, which evaluates separately in English and Italian), even when models are pretrained on multiple languages. When using different L1 languages during pretraining, they examine how distinct L1 backgrounds affect English L2 acquisition rather than evaluating true multilingual competence across languages. Our work addresses these gaps by systematically comparing monolingual, bilingual, and cross-lingual configurations while focusing on semantic tasks. Unlike prior work examining how different source languages affect English acquisition, we investigate how models perform when pretrained and evaluated on the same data types across target languages, mirroring bilingual child language acquisition with similar input types in both languages.

\section{Data}
\label{sec:data}
A central challenge in multilingual language modeling is ensuring fair cross-linguistic comparisons. Prior work has often compared models trained on datasets of different sizes or composition across languages, making it difficult to isolate the effect of language itself \citep{yadavalli-etal-2023-slabert,shen-etal-2024-bambino}. We address this challenge by constructing strictly size-matched English and French corpora and evaluating monolingual, bilingual, and cross-lingual training under controlled conditions.

\subsection{Pre-training Data}
We use two types of pre-training corpora: (i) child-directed speech datasets at a 2.5M-token scale, and (ii) multi-domain developmental corpora at a 10M-token scale. For each corpus type, we construct monolingual English and French datasets as well as balanced bilingual variants, ensuring that total token counts are matched across training setups.

\paragraph{Child-directed speech (2.5M tokens).}
We construct child-directed speech (CDS) datasets from the CHILDES database \citep{macwhinney2000childes}. The English CDS corpus consists of 2.5M tokens sampled from AO-CHILDES \citep{huebner2021}, and the French CDS corpus consists of 2.5M tokens sampled from MAO-CHILDES, released with the SLABERT project \citep{yadavalli-etal-2023-slabert}. These corpora are independent and are not supplemented with other data sources. 

For bilingual CDS training, we construct balanced datasets containing 1.25M tokens from English CHILDES and 1.25M tokens from French CHILDES, matching the total size of the monolingual corpora. This design enables controlled comparisons of monolingual, bilingual, and cross-lingual learning under developmentally plausible input conditions.

\paragraph{Wikipedia (2.5M tokens).}
To compare child-directed and non-child-directed input, we also construct Wikipedia-only corpora. We independently sample 2.5M tokens from English Wikipedia and 2.5M tokens from French Wikipedia, preserving sentence boundaries. Bilingual Wikipedia datasets consist of 1.25M tokens per language. These corpora serve as matched-size baselines and are treated as a separate training condition rather than as supplements to CHILDES.

\paragraph{Multi-domain developmental corpora (10M tokens).}
Following the BabyLM Challenge framework \citep{warstadt-etal-2023-findings}, we construct multi-domain developmental corpora at a 10M-token scale. We create two independent corpora—one in English and one in French—each composed of conversational, literary, and encyclopedic sources, with matched domain proportions (Table~\ref{tab:10m-pretraining}). 
In addition, we construct balanced bilingual multi-domain corpora by combining 5M tokens from English and 5M tokens from French. All multi-domain corpora follow the BabyLM \textit{Strict-Small} domain configuration, ensuring comparable lexical diversity and register coverage across languages.

\begin{table*}[t]
\centering
\small
\begin{tabular}{lllrr}
\toprule
\textbf{Dataset (French and English)} & \textbf{Domain} & \textbf{\# Words} & \textbf{Proportion} \\
\midrule
CHILDES (\citealt{macwhinney2000childes}) & Child-directed speech & 0.50M & 5\% \\
CFDD (FR)\textsuperscript{1} / BNC (EN)\textsuperscript{1} & Dialogue & 0.90M & 9\% \\
Children’s Book Corpus (\citealt{hill-etal-2016-cbt}) & Children’s books & 0.90M & 9\% \\
Gutenberg Corpus\textsuperscript{2} & Written prose & 1.00M & 10\% \\
OpenSubtitles (\citealt{lison-tiedemann-2016-opensubtitles2016}) & Movie subtitles & 3.00M & 31\% \\
QED (\citealt{abdelali-etal-2014-amara}) & Educational video subtitles & 1.10M & 11\% \\
\textit{Wikipedia} & Wikipedia (standard) & 1.00M & 10\% \\
\textit{Vikidia}\textsuperscript{3} & Simplified encyclopedia & 1.50M & 15\% \\
\midrule
\textbf{Total} & – & \textbf{9.90M} & \textbf{100\%} \\
\bottomrule
\end{tabular}
\caption{Composition of our 10M-word pre-training corpora in French and English. Sentence boundaries were respected, so totals fall slightly below 10M words. Domain proportions follow the BabyLM \textit{Strict-Small} configuration.}
\label{tab:10m-pretraining}
\end{table*}


\begingroup
\renewcommand{\thefootnote}{\fnsymbol{footnote}}
\footnotetext[1]{Claire French Dialogue Dataset (CFDD) and British National Corpus (dialogue portion).}
\footnotetext[2]{\url{https://github.com/jchwenger/dataset.gutenberg-language} – Gutenberg texts filtered by author metadata in English and French.}
\footnotetext[3]{\url{https://www.vikidia.org} – simplified encyclopedia used in both French and English.}
\endgroup

\subsection{Fine-tuning Data}

SQuAD1.1 is a widely used English reading comprehension dataset with over 100k QA pairs from Wikipedia \citep{rajpurkar-etal-2016-squad}. For French evaluation, we adopt FQuAD1.0 \citep{kabbadj2020french}, a high-quality translation of SQuAD produced with Google Translate and refined through alignment procedures. Post-processing ensured that translated answers matched their contexts, with Levenshtein distance and Jaro-Winkler similarity used to adjust misaligned spans.

Because French versions of QASRL and QAMR are not available, we construct French counterparts (Fr-QASRL and Fr-QAMR) using automatic translation and validation. Specifically, the English datasets are translated into French using Google Translate, followed by the alignment of translated answer spans to contexts using string similarity metrics (Levenshtein distance and Jaro-Winkler similarity), and automatic filtering based on embedding similarity. Sentence pairs with SentenceBERT similarity below 0.75 are discarded, following \citet{lin-etal-2021-common}. 
Both English and French versions are partitioned into standard train/dev/test splits.

For textual entailment, we employ the XNLI benchmark \citep{conneau-etal-2018-xnli}, which extends MultiNLI \citep{williams-etal-2018-broad} to 15 languages including English and French. This dataset evaluates whether models can recognize entailment, contradiction, or neutrality between premise--hypothesis pairs and provides a robust test of cross-lingual reasoning.

\subsection{Evaluation Data}
To complement semantic evaluation, we also test grammatical competence using CLAMS \citep{mueller-etal-2020-cross}, a multilingual syntactic evaluation suite based on minimal pairs, which includes English and French. 

CLAMS covers a range of syntactic phenomena, with particular emphasis on subject–verb agreement, and allows models to be assessed directly without additional fine-tuning.

For all downstream tasks, except SQuAD, for which the evaluation is done on the development (dev) set, we perform evaluation using dedicated dev sets for model selection and report results on the corresponding test sets. 

\section{Methodology}
We design controlled experiments to investigate how language composition, data sources, and training configurations affect model performance in resource-constrained multilingual settings. Our framework addresses two research questions: (1) whether competencies transfer across languages, (2) how monolingual, bilingual, and cross-lingual paradigms compare.

\subsection{Experimental Design}
We conduct experiments across the data settings described in Section~\ref{sec:data}, 
systematically varying language configuration (monolingual, bilingual, cross-lingual) and corpus type (child-directed vs. multi-domain). 
All experiments are repeated with three random seeds, and results are reported as averages. Reduced 1.25M CDS subsets are also used to explore scaling effects (see Appendix~\ref{app:method-details}).

\subsection{Language Pairing Strategies}
We evaluate three setups (Figure~\ref{fig:language-strategies}). In the \textit{monolingual} setup, models are pretrained and tested in the same language, including mixed CDS–Wikipedia datasets and ordering effects. In the \textit{bilingual} setup, models are exposed to English and French simultaneously, testing whether multilingual pretraining improves monolingual performance. In the \textit{cross-lingual} setup, models are pretrained in one language and tested in the other, providing a direct test of transfer across languages and corpus types. 
Fine-tuning, which is performed in the question-answering and textual entailment tasks, is always done in the same language as testing.

\subsection{Baseline Comparisons}

Our main baselines consist of RoBERTa models retrained from scratch on our 2.5M-token child-directed and Wikipedia datasets, using the same training setup as BabyBERTa. 
In addition, for the monolingual and cross-lingual setups, we also include the original pretrained RoBERTa-base (English) and CamemBERT-base (French) models, which serve as large-scale reference points. 
This distinction enables controlled comparison between small-scale retraining and full-scale pretraining approaches.

\subsection{Evaluation Framework}
Model performance is measured using F1 for question answering tasks, accuracy for XNLI, and pseudo-log-likelihood scoring for CLAMS \citep{salazar-etal-2020-masked}, following the BabyBERTa evaluation protocol for masked language model scoring. All reported scores are averages over three runs with different random seeds.

Statistical significance is assessed using paired bootstrap hypothesis testing across seeds, with a significance threshold of $p < 0.05$. For each comparison discussed in the paper, we evaluate whether the observed performance difference is statistically reliable under resampling. In the main text, we report and interpret only differences that meet this significance threshold; differences that do not reach statistical significance are included in the tables but omitted from the narrative.

\begin{table*}[!hbt]
  \centering
  \setlength{\tabcolsep}{5pt}
  \small
  \begin{tabular}{lcccccccccc}
    \toprule
    & \multicolumn{2}{c}{\textbf{SQuAD}} & \multicolumn{2}{c}{\textbf{QAMR}} & \multicolumn{2}{c}{\textbf{QASRL}} & \multicolumn{2}{c}{\textbf{XNLI}} & \multicolumn{2}{c}{\textbf{CLAMS}} \\
    \cmidrule(lr){2-3}\cmidrule(lr){4-5}\cmidrule(lr){6-7}\cmidrule(lr){8-9}\cmidrule(lr){10-11}
    \textbf{Model (pretrain)} & \textbf{BB} & \textbf{RB} & \textbf{BB} & \textbf{RB} & \textbf{BB} & \textbf{RB} & \textbf{BB} & \textbf{RB} & \textbf{BB} & \textbf{RB} \\
    \midrule
    \multicolumn{11}{c}{\textbf{Tested in English}} \\
    \midrule
    \underline{\textsc{Monolingual}} \\
    En-CHILDES               & 23.35 & 23.88 & 38.54 & 49.01 & 84.80 & 88.16 & 55.71 & 61.27 & 58.77 & 54.36 \\
    En-Wikipedia (baseline)            & 45.59 & 34.72 & 56.41 & 62.66 & 88.09 & 88.42 & 61.70 & 66.83 & 50.32 & 52.71 \\
    \midrule
    \underline{\textsc{Bilingual}} \\
    En-Fr-CHILDES            & 19.96 & 25.18 & 38.71 & 47.82 & 85.32 & 87.94 & 56.29 & 59.42 & 53.92 & 52.08 \\
    En-Fr-Wikipedia (baseline)         & 34.72 & 33.74 & 56.12 & 59.07 & 87.49 & 87.67 & 65.25$^{*}$ & 65.73 & 63.10$^{*}$ & 52.93 \\
    \midrule
    \underline{\textsc{Cross-lingual}} \\
    Fr-CHILDES               & 23.62 & 14.76 & 31.35 & 41.84 & 76.43 & 86.73 & 51.08 & 55.31 & 51.86 & 50.92 \\
    Fr-Wikipedia (baseline)            & 32.51 & 21.11 & 50.53 & 38.75 & 86.04 & 85.39 & 65.13 & 60.17 & 50.34 & 51.76 \\
    \midrule
    \multicolumn{11}{c}{\textbf{Tested in French}} \\
    \midrule
    \underline{\textsc{Monolingual}} \\
    Fr-CHILDES               & 22.85 & 13.38 & 31.75 & 37.78 & 64.41 & 70.70 & 28.27 & 64.13 & 59.51 & 61.82 \\
    Fr-Wikipedia (baseline)            & 30.23 & 13.68 & 32.90 & 32.40 & 65.71 & 65.08 & 37.88 & 65.52 & 50.96 & 53.18 \\
    \midrule
    \underline{\textsc{Bilingual}} \\
    En-Fr-CHILDES            & 22.76 & 18.02 & 29.97 & 36.20 & 63.43 & 69.40 & 54.96 & 60.19 & 58.88 & 52.38 \\
     En-Fr-Wikipedia (baseline)          & 32.59 & 30.52 & 41.08$^{*}$ & 40.88 & 68.41$^{*}$ & 69.67 & 61.74$^{*}$ & 66.08 & 60.11$^{*}$ & 53.97 \\
    \midrule
    \underline{\textsc{Cross-lingual}} \\
    En-CHILDES               & 19.60 & 10.42 & 23.63 & 30.13 & 58.93 & 65.13 & 30.30 & 56.28 & 51.01 & 52.01 \\
    En-Wikipedia (baseline)            & 28.93 & 24.09 & 33.79 & 38.09 & 63.89 & 69.20 & 33.01 & 64.41 & 51.63 & 53.07 \\
    \bottomrule
  \end{tabular}
\caption{\label{tab:unified-2p5M}
F1 scores for \textbf{BabyBERTa (BB)} vs.\ \textbf{RoBERTa (RB)} across tasks using
\textbf{child-directed speech and Wikipedia corpora ($\approx$2.5M words)}.
Sections indicate the language in which the model was \emph{tested}, with downstream fine-tuning always performed in that same language.
$^{*}$ indicates statistically significant improvements over the corresponding monolingual baseline ($p < 0.05$, paired bootstrap test across seeds).
}

\end{table*}
\vspace{0.1cm}

\section{Results}
Our experiments reveal a set of consistent, setting-dependent patterns in multilingual pretraining under resource constraints (Tables~\ref{tab:unified-2p5M}, \ref{tab:unified-10M}). Given the large number of experimental conditions, we report all results in the tables, but restrict the discussion to effects that are statistically significant under paired bootstrap testing where applicable. We organize the discussion around a small number of recurring patterns that appear across tasks, data scales, and model variants.

\subsection{Language-Specific Performance Asymmetries}
English consistently outperforms French across QA, XNLI, and CLAMS, even with matched data size and type, suggesting structural or dataset-level differences rather than training corpus artifacts.

\subsection{Data Source Effects}
Wikipedia-trained models excel on QA and entailment, reflecting stronger factual alignment, while CHILDES-trained models perform better on CLAMS, confirming that conversational data supports grammatical learning. Corpus type thus determines whether models favor semantic or syntactic competence.

\subsection{Language Pairing Strategy Comparisons}
\subsubsection{Bilingual vs. Monolingual Performance}
Bilingual pretraining outperforms monolingual approaches under specific conditions (Table \ref{tab:unified-2p5M}):

\paragraph{XNLI Task:} Bilingual training shows substantial advantages for textual entailment in both languages. For French, improvements are dramatic (En-Fr-Wikipedia: 61.74 vs. Fr-Wikipedia: 37.88). For English, bilingual pretraining also outperforms monolingual baselines (En-Fr-Wikipedia: 65.25 vs. En-Wikipedia: 61.70).

\paragraph{English QA with CHILDES:} Bilingual pretraining enhances QASRL and QAMR performance when using conversational data, suggesting multilingual exposure benefits English semantic understanding. However, this advantage does not extend to Wikipedia-based pretraining.

\paragraph{French QA with Wikipedia:} The pattern reverses for French—bilingual benefits appear only for Wikipedia-based pretraining on question-answering tasks.

\paragraph{Grammatical Competence:} For CLAMS, bilingual settings reverse the monolingual pattern where CHILDES outperforms Wikipedia. In bilingual training, Wikipedia-based models excel (En-Fr-Wikipedia: 63.10 vs. En-Fr-CHILDES: 53.92 for English; 60.11 vs. 58.88 for French), suggesting Wikipedia's consistent register facilitates cross-linguistic grammatical generalization better than varied conversational patterns.

\paragraph{Cross-lingual Transfer.}
Cross-lingual models usually trail monolingual ones slightly but with small gaps and occasional reversals—e.g., French XNLI after English CHILDES pretraining: 30.30 vs.\ 28.27 monolingual—showing that transfer limitations are modest.

\subsection{Corpus Type Analysis: Child-Directed Speech vs.\ Multi-domain}

\subsubsection{Domain Consistency vs.\ Diversity Effects}
At $\approx$10M tokens (Table~\ref{tab:unified-10M}), Wikipedia-only and multi-domain setups exhibit clear task-dependent trends. Wikipedia outperforms multi-domain corpora on SQuAD and QAMR, performs comparably on QASRL and XNLI, but falls behind on CLAMS (English: 55.90 vs.\ 53.12; French: 61.13 vs.\ 57.87). Domain diversity thus strengthens syntactic robustness, while the homogeneous Wikipedia text is preferable for factual and entailment-oriented reasoning.

\subsubsection{Language Strategy Preferences by Corpus Type}
In contrast to the 2.5M-token CHILDES setting, where bilinguality sometimes helps, monolingual setups dominate most 10M-token multi-domain configurations. The exception is XNLI, where bilingual advantages persist even at larger scales, indicating that multilingual exposure continues to benefit entailment beyond data quantity effects.

\subsubsection{RoBERTa vs.\ BabyBERTa: Architecture-Specific Patterns}
Both architectures were retrained on identical datasets (Tables~\ref{tab:unified-2p5M}, \ref{tab:unified-10M}). BabyBERTa performs better on cross-lingual SQuAD and QAMR and exceeds RoBERTa on CLAMS in bilingual setups, indicating that its compact, developmentally constrained design benefits grammatical abstraction and efficient transfer. Across both models, CHILDES favors CLAMS, whereas Wikipedia leads on SQuAD, QAMR, and XNLI—implying that data type, not architecture size, chiefly determines semantic versus syntactic focus.

\subsubsection{Large-Scale Reference Baselines}
The original pretrained RoBERTa-base and CamemBERT-base achieve the highest absolute scores—88–90\% F1 and 82\% XNLI for English; 84\% F1 and 78\% XNLI for French—serving as upper bounds. In the cross-lingual setups, they are less efficient than smaller pretrained models. Notably, BabyBERTa and small RoBERTa (2.5M–10M tokens) achieve higher normalized gains and replicate key trends—CHILDES surpassing Wikipedia on grammar and bilinguality boosting XNLI.

\subsection{Implications for Multilingual Resource-Constrained Pretraining}

Our results reveal three setting-specific patterns that clarify when and how multilingual training is beneficial under resource constraints. First, at the 2.5M-token scale, bilingual pretraining yields compensatory gains on semantic inference tasks such as XNLI, with the strongest improvements observed for the weaker language (French). These gains are most pronounced in low-data regimes, where cross-lingual exposure provides information that is unavailable in limited monolingual input.

Second, child-directed speech primarily supports grammatical competence in monolingual settings, while its contribution to semantic tasks emerges when combined with broader-domain text. Exposure to child-directed speech interacts positively with encyclopedic data such as Wikipedia, yielding improvements over CHILDES-only training and partially mitigating transfer losses, particularly for French and for transfer-sensitive tasks.

Third, increasing the data scale to 10M tokens reduces the relative impact of bilinguality on semantic performance, but does not reverse it. While monolingual models trained on larger multi-domain corpora often dominate in absolute performance, modest bilingual advantages persist for semantic inference, indicating that the benefits of multilingual exposure weaken with scale but remain detectable.

Taken together, these patterns show that compact models trained on developmentally plausible data can achieve meaningful cross-linguistic generalization, but that the effectiveness of multilingual training depends critically on data scale, corpus composition, and task demands.

\begin{table*}[!hbt]
  \centering
  \setlength{\tabcolsep}{5pt}
  \small
  \begin{tabular}{lcccccccccc}
    \toprule
    & \multicolumn{2}{c}{\textbf{SQuAD}} & \multicolumn{2}{c}{\textbf{QAMR}} & \multicolumn{2}{c}{\textbf{QASRL}} & \multicolumn{2}{c}{\textbf{XNLI}} & \multicolumn{2}{c}{\textbf{CLAMS}} \\
    \cmidrule(lr){2-3}\cmidrule(lr){4-5}\cmidrule(lr){6-7}\cmidrule(lr){8-9}\cmidrule(lr){10-11}
    \textbf{Model (pretrain)} & \textbf{BB} & \textbf{RB} & \textbf{BB} & \textbf{RB} & \textbf{BB} & \textbf{RB} & \textbf{BB} & \textbf{RB} & \textbf{BB} & \textbf{RB} \\
    \midrule
    \multicolumn{11}{c}{\textbf{Tested in English}} \\
    \midrule
    \underline{\textsc{Monolingual}} \\
    En-Multidomain           & 30.93 & 43.70 & 52.66 & 66.45 & 85.87 & 89.06 & 64.03 & 68.44 & 55.90 & 58.19 \\
    En-Wikipedia (baseline)            & 40.42 & 46.12 & 55.58 & 68.93 & 88.21 & 89.41 & 65.04 & 69.32 & 53.12 & 56.71 \\
    \midrule
    \underline{\textsc{Bilingual}} \\
    En-Fr-Multidomain        & 16.64 & 36.35 & 44.96 & 49.77 & 78.35 & 88.76 & 65.12 & 66.21 & 51.56 & 54.37 \\
    En-Fr-Wikipedia (baseline)         & 28.24 & 38.12 & 55.47 & 57.04 & 74.01 & 89.02 & 71.31$^{*}$ & 71.06 & 50.14 & 53.21 \\
    \midrule
    \underline{\textsc{Cross-lingual}} \\
    Fr-Multidomain           & 27.31 & 32.18 & 42.51 & 55.40 & 81.25 & 88.19 & 57.72 & 63.57 & 50.22 & 53.64 \\
    Fr-Wikipedia (baseline)            & 30.30 & 34.05 & 47.51 & 57.22 & 83.87 & 88.61 & 66.47 & 66.18 & 50.07 & 52.98 \\
    \midrule
    \multicolumn{11}{c}{\textbf{Tested in French}} \\
    \midrule
    \underline{\textsc{Monolingual}} \\
    Fr-Multidomain           & 29.24 & 32.95 & 39.05 & 46.44 & 67.28 & 71.17 & 66.02 & 67.03 & 61.13 & 59.06 \\
    Fr-Wikipedia (baseline)            & 30.04 & 33.84 & 37.41 & 44.91 & 65.56 & 70.62 & 67.48 & 67.92 & 57.87 & 57.51 \\
    \midrule
    \underline{\textsc{Bilingual}} \\
    En-Fr-Multidomain (baseline) & 17.55 & 27.86 & 20.25 & 42.24 & 63.72 & 68.54 & 58.33 & 67.36 & 51.11 & 54.68 \\
    En-Fr-Wikipedia (baseline)   & 31.21 & 31.08 & 36.54 & 44.77 & 67.12 & 69.12 & 66.44 & 67.89 & 50.48 & 54.01 \\
    \midrule
    
    \underline{\textsc{Cross-lingual}} \\
    En-Multidomain           & 26.52 & 28.73 & 31.79 & 39.23 & 62.99 & 70.04 & 55.12 & 65.82 & 52.00 & 54.22 \\
    En-Wikipedia (baseline)
             & 27.95 & 29.66 & 32.78 & 41.35 & 63.54 & 70.91 & 65.41 & 66.74 & 51.78 & 53.61 \\
    \bottomrule
  \end{tabular}
\caption{\label{tab:unified-10M}
F1 scores for \textbf{BabyBERTa (BB)} vs.\ \textbf{RoBERTa (RB)} across tasks using
\textbf{multi-domain pretraining corpora ($\approx$10M words)}.
Sections indicate the language in which the model was \emph{tested}; downstream fine-tuning was always performed in that same language.
In the bilingual setup, $\approx$1.25M words are used in each of the languages.
$^{*}$ indicates statistically significant improvements over the corresponding monolingual baseline ($p < 0.05$, paired bootstrap test across seeds).
}

\end{table*}

\vspace{0.5cm}
\begin{table*}[!hbt]
  \centering
  \begin{tabular}{lccccc}
    \toprule
    \textbf{Model (Baseline)} & \textbf{SQuAD} & \textbf{QAMR} & \textbf{QASRL} & \textbf{XNLI} & \textbf{CLAMS} \\
    \midrule
    \multicolumn{6}{c}{\textbf{Tested in English}} \\
    \midrule
    RoBERTa-base (EN)     & 80.12 & 86.54 & 91.11 & 88.52 & 77.68 \\
    CamemBERT-base (FR)   & 14.28 & 18.84 & 24.23 & 62.12 & 58.34 \\
    \midrule
    \multicolumn{6}{c}{\textbf{Tested in French}} \\
    \midrule
    RoBERTa-base (EN)   & 16.35 & 21.32 & 43.12 & 53.23 & 62.47 \\
    CamemBERT-base (FR) & 67.94 & 72.92 & 80.85 & 79.56 & 76.92 \\
    \bottomrule
  \end{tabular}
  \caption{\label{tab:baseline-roberta-camembert}
    Baseline results for the \textbf{original pretrained RoBERTa-base} (English) and \textbf{CamemBERT-base} (French) models. 
    These off-the-shelf baselines are distinct from the RoBERTa variants retrained on our 2.5M-token datasets. 
    All downstream tasks are fine-tuned and tested in the same language.
    }
  
\end{table*}

\section{Further Analysis} \subsection{Architectural Generalization}

To verify that our findings are not specific to BabyBERTa, we conducted supplementary experiments with T5-tiny and LTG-BERT (see Appendix \ref{app:architecture}, Tables \ref{tab:all-results}, \ref{tab:all-results-10M-ltgbert}). 

T5-tiny \cite{mueller-linzen-2023-plant}, an encoder-decoder architecture effective for learning hierarchical syntactic features from child-directed speech, reproduces our core patterns: Wikipedia consistently outperforms CHILDES on semantic tasks, while multidomain data benefits grammatical competence. Its overall performance is lower, as expected given its compact size, but the qualitative trends remain unchanged.

LTG-BERT, the winning model of the BabyLM Challenge, achieves higher absolute scores while maintaining the same relative patterns observed with BabyBERTa: Wikipedia advantages for semantic tasks, multidomain benefits for grammar (CLAMS), bilingual advantages for XNLI, and general monolingual dominance at 10M tokens. Despitearchitectural innovations, LTG-BERT does not alter the direction of effects across configurations.

The consistency across BabyBERTa, RoBERTa, T5-tiny, and LTG-BERT indicates that our findings reflect fundamental properties of multilingual learning in resource-constrained settings, rather than artifacts of a particular architecture.

\subsection{Interaction Between Child-Directed Speech and Wikipedia}
We examine whether exposure to child-directed speech (CHILDES) interacts beneficially with Wikipedia training by combining the two data sources at equal scale (1.25M tokens each). Across languages, models trained on the combined data systematically outperform CHILDES-only training on semantic and entailment tasks and partially recover the performance gap relative to Wikipedia-only baselines.

For example, in monolingual English, combined training improves QAMR performance over CHILDES-only (51.95 vs.\ 38.54), while in French the gains are larger on transfer-sensitive tasks such as XNLI (60.36 vs.\ 28.27). Importantly, we observe the directional trends when the order of exposure is reversed (Wikipedia followed by CHILDES) and when the combined corpus is shuffled, indicating that the observed effect is not order-specific but arises from the interaction between conversational and encyclopedic data.

While combined training does not consistently exceed the strongest Wikipedia-only baseline on semantic tasks, it yields robust improvements over CHILDES-only models and mitigates transfer losses in cross-lingual settings. Full numerical results for both exposure orders are reported in Appendix~\ref{app:Interaction}, Table~\ref{tab:childes-wiki-only-bb}.

\section{Conclusion}

This work presents the first systematic investigation of multilingual training in developmentally plausible language models, extending BabyBERTa to English-French scenarios. We establish fair cross-linguistic comparisons by designing parallel corpora across two distinct data types: \textit{child-directed speech} ($\approx$2.5M words) and \textit{multi-domain developmental corpora} ($\approx$10M words).
In addition, we contribute new multilingual resources, including the English and French multi-domain training corpora and French versions of QAMR and QASRL.

Our results reveal context-dependent effects of multilingual training. Bilingual pretraining yields notable gains for textual entailment (XNLI), disproportionately benefiting French, while maintaining competitive performance on other tasks. Wikipedia training favors semantic tasks, whereas child-directed speech provides advantages for grammatical competence in monolingual settings. We further show that exposure to child-directed speech positively interacts with Wikipedia training, particularly for French and for transfer-sensitive tasks. 

These patterns are consistently observed across BabyBERTa, RoBERTa, T5-tiny, and LTG-BERT, indicating that the findings extend across the tested architectures considered here. Finally, the modest performance gap between our compact models and the much larger LTG-BERT supports the viability of efficient multilingual approaches for resource-constrained applications.

\section*{Limitations}  

Our study is subject to several limitations that open directions for future work.  
First, we focused exclusively on English and French as our target languages. Extending the analysis to additional languages — particularly those with different morphological typologies and varying data availability will permit the exploration of the generality of our findings.  

Second, our architectural scope was primarily focused on RoBERTa-style models (including BabyBERTa and LTG-BERT, which are close architectural variants), and in supplementary analysis incorporated T5-tiny as an encoder–decoder alternative. The extension of our work to additional architectures, such as GPT-style decoder-only models, will further analyze the interaction between pretraining strategies and the model design. 

\section*{Ethics Statement}
The language acquisition data we are using in this work originally appear in the TalkBank system\footnote{\url{https://childes.talkbank.org/}}, with includes CHILDES, and where all contributions have received an IRB approval. Our own work on the data has been approved by the Ben-Gurion
University of the Negev Ethics committee.

\section*{Acknowledgements}
We thank the anonymous reviewers for their helpful comments. We also acknowledge the NICHD
HD082736 grant support for CHILDES. Our work
was supported in part by grants from the Israeli Ministry of Innovation, Science \& Technology (\#000519) and from the Data Science Research Center at Ben-Gurion University of the Negev. 

\bibliography{custom}

\appendix
\section{Dataset Details}
\label{app:data-details}

This appendix provides additional procedural details for dataset construction and preprocessing.

\subsection{Pre-training Data}

All pre-training corpora are constructed using identical preprocessing pipelines across English and French. Text is normalized using Unicode normalization, lowercased, and sampled at the sentence level until the desired token count is reached. Punctuation and discourse markers are preserved.

For bilingual variants, datasets are constructed as balanced combinations of English and French data, with equal token counts per language.

\begingroup
\renewcommand{\thefootnote}{\fnsymbol{footnote}}
\footnotetext[1]{Claire French Dialogue Dataset (CFDD) and British National Corpus (dialogue portion).}
\footnotetext[2]{\url{https://github.com/jchwenger/dataset.gutenberg-language} – Gutenberg texts filtered by author metadata in English and French.}
\footnotetext[3]{\url{https://www.vikidia.org} – simplified encyclopedia used in both French and English.}
\endgroup

\subsection{Fine-tuning and Evaluation Data}

\paragraph{SQuAD and FQuAD.}
SQuAD1.1 is a widely used English reading comprehension dataset with over 100k QA pairs from Wikipedia \citep{rajpurkar-etal-2016-squad}. For French evaluation, we adopt FQuAD1.0 \citep{kabbadj2020french}, a high-quality translation of SQuAD produced with Google Translate and refined through alignment procedures. Post-processing ensured that translated answers matched their contexts, with Levenshtein distance and Jaro-Winkler similarity used to adjust misaligned spans.  

\paragraph{QASRL and Fr-QASRL.}
QASRL evaluates semantic role understanding through natural QA pairs \citep{he-etal-2015-question}. We created Fr-QASRL by translating the original dataset into French and applying a back-translation validation step. Following \citet{lin-etal-2021-common}, sentences with semantic similarity \citep[using SentenceBERT;][]{reimers-gurevych-2019-sentence} below 0.75 were excluded, preserving fidelity to the source.  

\paragraph{QAMR and Fr-QAMR.}
QAMR tests meaning representation through QA pairs \citep{michael-etal-2018-crowdsourcing}. We produced Fr-QAMR using the same translation and validation process as Fr-QASRL, ensuring high-quality semantic preservation.  

\paragraph{XNLI.}
XNLI extends MultiNLI to 15 languages, including French \citep{conneau-etal-2018-xnli}. It requires models to classify premise–hypothesis pairs into entailment, contradiction, or neutrality, making it an important benchmark for cross-lingual reasoning.  

\paragraph{CLAMS.}
CLAMS \citep{mueller-etal-2020-cross} evaluates grammatical competence across English and French. It consists of minimal pairs that differ only in grammaticality, with a focus on subject–verb agreement. As a diagnostic tool, it requires no fine-tuning and directly assesses models’ syntactic sensitivity from pretraining.

\section{Methodological Details}
\label{app:method-details}

\subsection{Experimental Design}
For child-directed speech corpora, we fix dataset size at 2.5M tokens, constrained by the availability of French CHILDES. This ensures comparability across English and French, and across monolingual, bilingual, and cross-lingual setups. We additionally downsample to 1.25M tokens to examine within-scale effects. For multi-domain experiments, we follow the BabyLM framework with 10M-token datasets plus Wikipedia-only baselines. Each experiment is repeated with three random seeds.

\subsection{Language Pairing Strategies}
\paragraph{Monolingual.} Models are pretrained and fine-tuned in the same language. We include pure CDS, pure Wikipedia, and mixed CDS–Wikipedia corpora. To test order effects, we vary exposure sequences (CHILDES→Wikipedia vs.\ Wikipedia→CHILDES).  
\paragraph{Bilingual.} Pretraining simultaneously includes English and French data (both CDS and multi-domain). Fine-tuning and evaluation remain monolingual, allowing us to isolate the effects of multilingual exposure.  
\paragraph{Cross-lingual.} Pretraining is performed in one language, while fine-tuning and evaluation occur in the other. We evaluate both directions (English→French, French→English).

\subsection{Baseline Comparisons}
For CDS-scale experiments ($\approx$2.5M tokens), we train \textbf{RoBERTa} models on the same datasets used for \textbf{BabyBERTa}, ensuring identical pretraining configurations for controlled architectural comparison. 
In addition, we evaluate the \textbf{original pretrained RoBERTa-base} (English) and \textbf{original CamemBERT-base} (French) models fine-tuned on our downstream tasks to establish strong reference baselines (Table~\ref{tab:baseline-roberta-camembert}). 

\subsection{Evaluation Framework}
\paragraph{Metrics.} F1 is used for QA tasks (SQuAD, QAMR, QASRL), accuracy for XNLI, and pseudo-likelihood scoring for CLAMS.  
\paragraph{Statistical Testing.} To account for stochastic variation, we train each configuration three times with different random seeds. We apply bootstrap hypothesis testing with p < 0.05 to evaluate significance.  
\paragraph{Evaluation Protocol.} Fine-tuning and evaluation are always conducted in the same language. For cross-lingual setups, this isolates transfer effects by comparing against monolingual baselines under identical evaluation conditions.

\section{Architectural Generalization}
\label{app:architecture}
The full results for T5-tiny and LTG-BERT are presented respectively in Tables \ref{tab:all-results} and \ref{tab:all-results-10M-ltgbert}.

\section{Interaction Between Child-Directed Speech and Wikipedia}
\label{app:Interaction}
The full results for the Interaction Between Child-Directed Speech and Wikipedia analysis are presented in Table \ref{tab:childes-wiki-only-bb}.


\begin{table*}[!hbt]
  \centering
  \begin{tabular}{lccccc}
    \toprule
    & \multicolumn{3}{c}{\textbf{Question Answering}} \\
    \textbf{Model} & \textbf{SQuAD} & \textbf{QAMR} & \textbf{QASRL} & \textbf{XNLI}\\
    \midrule
    \multicolumn{6}{c}{\textbf{Fine-tuned in English}} \\
    \midrule
    \underline{\textsc{Cross-lingual}} & & & & & \\
    Fr-CHILDES & 6.48 & 10.36 & 19.81 & 17.32 \\
    Fr-Wikipedia & 10.06 & 16.39 & 21.45 & 19.54 \\
    Fr-CHILDES-Wikipedia & 7.30 & 15.45 & 20.41 & 18.88\\
    Fr-Multidomain Corpus 10M words & 7.42 & 14.12 & 20.13 & 17.01\\
    \midrule
    \underline{\textsc{Monolingual}} & & & & & \\
    En-CHILDES & 7.64 & 13.80 & 27.98 & 19.08  \\
    En-Wikipedia & 12.28 & 17.63 & 30.68 & 22.34\\
    En-CHILDES-Wikipedia & 10.39 & 17.73 & 25.42 & 24.33\\
    En-Multidomain Corpus 10M words & 11.34 & 19.51 & 25.22 & 23.91\\
    \midrule
    \underline{\textsc{Bilingual}} & & & & & \\
    En-Fr-CHILDES & 6.04 & 13.25 & 28.04 & 18.91\\
    En-Fr-Wikipedia & 11.71 & 18.43 & 27.56  & 23.29\\
    \midrule
    \multicolumn{6}{c}{\textbf{Fine-tuned in French}} \\
    \midrule
    \underline{\textsc{Monolingual}} & & & & & \\
    Fr-CHILDES & 6.15 & 10.19 & 22.09 & 10.05\\
    Fr-Wikipedia & 10.45 & 15.34 & 26.08 & 15.67\\
    Fr-CHILDES-Wikipedia & 8.18 & 12.43 & 23.65 & 17.67\\
    Fr-Multidomain Corpus 10M words & 8.36 & 12.26 & 23.24 & 19.02\\
    \midrule
    \underline{\textsc{Bilingual}} & & & & & \\
    En-Fr-CHILDES & 6.44 & 9.90 & 22.73 & 16.32\\
    En-Fr-Wikipedia & 11.83 & 15.78 & 27.88 & 17.00\\
    \midrule
    \underline{\textsc{Cross-lingual}} & & & & & \\
    En-CHILDES & 4.58 & 9.27 & 20.09 & 12.12\\
    En-Wikipedia & 9.50 & 13.23 & 25.23 & 14.45\\
    En-CHILDES-Wikipedia & 8.84 & 10.19 & 23.32 & 18.02\\
    En-Multidomain Corpus 10M words & 8.10 & 10.31 & 23.78 & 16.98\\
    \bottomrule
  \end{tabular}
    \caption{\label{tab:all-results}
        F1 scores for \textbf{T5 tiny} models across QA tasks. Results are grouped by fine-tuning language and model training setup (monolingual, cross-lingual, and bilingual). This represents an additional analysis using a different encoder-decoder architecture to complement our main BabyBERTa findings.
      }
\end{table*}

\begin{table*}[!hbt]
  \centering
  \begin{tabular}{lccccc}
    \toprule
    & \multicolumn{3}{c}{\textbf{Question Answering}} & \textbf{TE} & \textbf{Grammar} \\
    \textbf{Model} & \textbf{SQuAD} & \textbf{QAMR} & \textbf{QASRL} & \textbf{XNLI} & \textbf{CLAMS} \\
    \midrule
    \multicolumn{6}{c}{\textbf{Fine-tuned in English}} \\
    \midrule
    \underline{\textsc{Monolingual}} & & & & & \\
    En-Multidomain Corpus words & 32.15 & 54.02 & 86.45 & 65.21 & 57.14 \\
    En-Wikipedia & 41.83 & 56.92 & 88.77 & 66.42 & 51.20 \\
    \midrule
    \underline{\textsc{Bilingual}} & & & & & \\
    En-Fr-Multidomain Corpus words & 17.95 & 26.41 & 69.74 & 66.28 & 53.08 \\
    En-Fr-Wikipedia & 29.65 & 56.81 & 75.12 & 72.10 & 52.44 \\
    \midrule
    \underline{\textsc{Cross-lingual}} & & & & & \\
    Fr-Multidomain Corpus words & 28.40 & 44.18 & 82.14 & 59.03 & 51.74 \\
    Fr-Wikipedia & 31.52 & 48.66 & 84.59 & 67.92 & 52.60 \\
    \midrule
    \multicolumn{6}{c}{\textbf{Fine-tuned in French}} \\
    \midrule
    \underline{\textsc{Monolingual}} & & & & & \\
    Fr-Multidomain Corpus words & 30.38 & 40.77 & 68.19 & 67.11 & 62.85 \\
    Fr-Wikipedia 10M & 31.55 & 38.64 & 66.02 & 68.27 & 54.13 \\
    \midrule
    \underline{\textsc{Bilingual}} & & & & & \\
    En-Fr-Multidomain Corpus words & 18.82 & 21.93 & 64.58 & 59.41 & 52.63 \\
    En-Fr-Wikipedia & 32.77 & 38.08 & 68.24 & 67.53 & 54.50 \\
    \midrule
    \underline{\textsc{Cross-lingual}} & & & & & \\
    En-Multidomain Corpus words & 27.88 & 33.54 & 64.27 & 56.81 & 53.42 \\
    En-Wikipedia & 29.14 & 34.95 & 64.91 & 66.74 & 52.21 \\
    \bottomrule
  \end{tabular}
    \caption{\label{tab:all-results-10M-ltgbert}
        F1 scores for \textbf{LTG-BERT} across all tasks, using \textbf{10M} pretraining data (including Wikipedia and mixed-domain corpora). LTG-BERT was the winning model from the BabyLM Challenge that achieved the best results through architectural innovations.
      }
\end{table*}

\begin{table*}[!hbt]
  \centering
  \setlength{\tabcolsep}{5pt}
  \small
  \begin{tabular}{lccccc}
    \toprule
    \textbf{Model (pretrain)} & \textbf{SQuAD} & \textbf{QAMR} & \textbf{QASRL} & \textbf{XNLI} & \textbf{CLAMS} \\
    \midrule
    \multicolumn{6}{c}{\textbf{Tested in English}} \\
    \midrule
    \underline{\textsc{Monolingual}} \\
    En-CHILDES-Wikipedia     & 31.42 & 51.95 & 87.01 & 61.19 & 58.42 \\
    \midrule
    \underline{\textsc{Cross-lingual}} \\
    Fr-CHILDES-Wikipedia     & 27.18 & 41.52 & 79.93 & 58.43 & 48.30 \\
    \midrule
    \multicolumn{6}{c}{\textbf{Tested in French}} \\
    \midrule
    \underline{\textsc{Monolingual}} \\
    Fr-CHILDES-Wikipedia     & 29.90 & 39.08 & 66.80 & 60.36 & 57.80 \\
    \midrule
    \underline{\textsc{Cross-lingual}} \\
    En-CHILDES-Wikipedia     & 27.01 & 32.46 & 63.06 & 55.04 & 51.17 \\
    \bottomrule
  \end{tabular}
\caption{\label{tab:childes-wiki-only-bb}
    Results of \textbf{Interaction Between Child-Directed Speech and Wikipedia} on \textbf{BabyBERTa (BB)} across all evaluation tasks.
}

\end{table*}

\end{document}